\documentclass{article}
\usepackage{graphicx} 
\usepackage{authblk}
\usepackage{graphicx}
\usepackage{hyperref}
\usepackage{xcolor}
\usepackage{xspace}
\usepackage{subcaption}
\usepackage{multirow}
\usepackage{comment}
\usepackage{dsfont}
\usepackage{enumitem}

\makeatletter
\DeclareRobustCommand\onedot{\futurelet\@let@token\@onedot}
\def\@onedot{\ifx\@let@token.\else.\null\fi\xspace}

\def\eg{\emph{e.g}\onedot}

\def\etal{\emph{et al}\onedot}
\makeatother

\title{Autoproof: Automated Segmentation Proofreading for Connectomics}
\author[1]{Gary \nolinebreak B \nolinebreak Huang}
\author[1]{William \nolinebreak M \nolinebreak Katz}
\author[1]{Stuart \nolinebreak Berg}
\author[1]{Louis \nolinebreak Scheffer}
\affil[1]{Janelia Research Campus, Howard Hughes Medical Institute, USA}
\date{\today}

\begin{document}

\maketitle

\begin{abstract}
Producing connectomes from electron microscopy (EM) images has historically required a great deal of human proofreading effort.  This manual annotation cost is the current bottleneck in scaling EM connectomics, for example, in making larger connectome reconstructions feasible, or in enabling comparative connectomics where multiple related reconstructions are produced.  In this work, we propose using the available ground-truth data generated by this manual annotation effort to learn a machine learning model to automate or optimize parts of the required proofreading workflows.  We validate our approach on a recent complete reconstruction of the \emph{Drosophila} male central nervous system.  We first show our method would allow for obtaining 90\% of the value of a guided proofreading workflow while reducing required cost by 80\%.  We then demonstrate a second application for automatically merging many segmentation fragments to proofread neurons.  Our system is able to automatically attach 200 thousand fragments, equivalent to four proofreader years of manual work, and increasing the connectivity completion rate of the connectome by 1.3\% points.  
\end{abstract}

\section{Introduction}

Significant progress has been made in the field of electron microscopy (EM) connectomics, with recent years seeing the release of large-scale connectome reconstructions~\cite{scheffer2020connectome,winding2023connectome,Takemura2024,dorkenwald2024neuronal}.  The primary bottleneck in producing such reconstructions is the amount of manual proofreading effort that must be applied to the automated segmentation.  This is a consequence of the need for very high accuracy in the segmentation, due to the potential impact of a false merge or split on the quality of the resulting connectome.  This manual proofreading cost has been estimated to be on the order of 50-100 proofreading years of effort~\cite{scheffer2020connectome}.  Reducing this manual annotation bottleneck is a key component toward scaling connectome reconstructions, such as toward making larger reconstructions feasible, or enabling comparative studies where multiple related reconstructions are produced.

Although much manual proofreading effort is applied to produce a reconstruction, the ground-truth data contained in these manual annotations is not typically used in the machine learning models used to produce the automated segmentations.  Generally, some manual review is done to check and validate a model, at which point the model's segmentation is used as a starting point for manual proofreading, using the workflows described in Section~\ref{sec:workflows}.  Our core idea is that there is valuable information contained in these ground-truth annotations, that can be used to train a model to help automate or optimize parts of the manual proofreading work.  We envision such a system, which we term \emph{Autoproof}, being deployed in a ``in-the-loop'' or active learning framework, being iteratively re-trained with newly available ground-truth from manual proofreading work, and whose output is iteratively used to automate or semi-automate such work.

In this paper, we describe the details of this system for automating segmentation proofreading.  We validate our approach on data from a large-scale complete reconstruction of \emph{Drosophila} male central nervous system, and apply our method to automate parts of the proofreading of the reconstruction.  Our contributions include:

\begin{itemize}[nosep]
\item A model for the focused proofreading workflow that incorporates grayscale, segmentation, shape, and synaptic information, achieving high accuracy and allowing for 90\% of the merges to be obtained with 20\% of the effort with no introduction of errors.
\item A system for applying the model from focused proofreading to the orphan link workflow, applied to a million unexamined orphans and generating 200 thousand proposed merge with sufficient accuracy to be automatically accepted into the final reconstruction.  These automated merges added 309 thousand T-bars and 2.4 million PSDs to the reconstruction, increasing the connectivity completeness rate by 1.3\% points, and constituted the equivalent of 4 person-years of manual proofreading effort.
\end{itemize}

\section{Background}

In this section, we first review the proofreading workflows used to produce large-scale EM connectome reconstructions, and then discuss work related to automated proofreading.

\subsection{Connectomics Segmentation Workflows}
\label{sec:workflows}

Given the scale of EM image data, and accuracy of current automated segmentation methods, ensuring the correctness of an entire segmentation is intractable.  Instead, manual proofreading effort is generally applied in one of several defined workflows, with the goal of achieving a certain level of connectome completeness.  Completeness is often measured in terms of connectivity completeness, where segments are divided into two groups: segments that have been sufficiently reconstructed such that the underlying neuron can be identified, and all other remaining fragments.  Connectivity completeness is then the percentage of all synapses where both pre- and post-synaptic partners belong to identified neurons~\cite{Takemura2024}.

In proofreading a segmentation, there are three primary workflows that are applied~\cite{scheffer2020connectome}: cleaving, focused proofreading, and orphan linking.  Cleaving is an initial step used to correct for false merges in the initial segmentation, while focused proofreading and orphan linking are workflows used to correct for false splits.  In terms of manual effort required, addressing false merges is generally more costly than addressing false splits, therefore segmentation models are often tuned to produce more conservative segmentations that minimize false merges, with a trade-off of more false splits.

Focused proofreading is a guided workflow for correction of false splits.  A set of candidate merge corrections is generated, consisting of a pair of segments and the spatial location where they may be merged.  For each candidate, a proofreader decides whether to accept the proposed merge or not.  The set of candidate merge corrections can be mined from a segmentation model; in practice the candidate merges are segment pairs that the segmentation model assigned an intermediate score for being merged, somewhat below the threshold needed to accept as an automatic merge.

Orphan link is an unguided workflow, where a proofreader is presented with a disconnected orphan fragment, with the goal of finding merges to connect the fragment back to a larger (identified) segment.  As the proofreader needs to consider the various locations where the fragment may possibly be merged, this task is more time-consuming than focused proofreading.  Generally focused proofreading is first applied until the set of likely merge candidates is fully examined.  Afterward, orphan link is applied to the remaining disconnected fragments, in decreasing order of synapse weight (synapses contained within the fragment).

\subsection{Related Work}

Other work in connectomics has considered the problem of automating proofreading.  NEURD presents a method for automatic correction of merge errors~\cite{celii2025neurd}.  This approach is complementary to our method, as NEURD relies on domain knowledge and hand-crafted features, whereas our method relies on learning models for automatic proofreading directly from manual annotations.  

Our method is also similar to Zung~\etal~\cite{zung2017error}, who propose a framework for segmentation error detection and correction.  Our work is distinct in incorporating shape and synapse-level information, being directly linked to the manual proofreading workflows used for large-scale connectomic reconstructions, and both validating our approach on such data as well as using our system to generate merges that were automatically incorporated into the reconstruction.

A method for automated proofreading is also given by Chen~\etal~\cite{chen2024learning}.  Again, our work is distinct in that our method is directly tested and evaluated within the manual proofreading workflows used for producing the reconstructions, and used to assist in the reconstruct, rather than constructing a separate benchmark for evaluation that may not exactly reflect the distribution of errors encountered during manual proofreading.

Finally, an alternative approach to automated proofreading is given by the method of RoboEM~\cite{schmidt2024roboem}, where a CNN model is trained to steer through an EM volume and avoid membranes.  This method showed an improvement to segmentation of mammalian tissue; we show in this paper similar improvements for invertebrate reconstruction while also being tightly coupled to the reconstruction proofreading workflows.

\section{Methods}

The core component of Autoproof is a convolutional network trained on focused merge decisions.  We first describe this model, then describe additional sources of information we have also considered.

\subsection{CNN for Focused Merge}

The task in the focused merge workflow is naturally formulated as a machine learning problem: given the inputs of the two candidate segments and a corresponding spatial location, the task corresponds to assigning a binary label of ``merge'' or ``don't merge'', depending on whether the two segments should be joined based on the location information at the given spatial location.

To make the binary prediction, we use a 3D convolutional neural network (CNN) with a VGG-like design of convolutional layers followed by fully connected layers~\cite{simonyan2014very}.  The network is given as input the local grayscale centered at the provided spatial location, and trained using the binary cross-entropy loss.  However, the model also needs to be aware of the proposed segments to be merged.  Therefore, we add in two additional 3D channels of information corresponding to binary masks of the local segmentation for each segment.  More precisely, given the local segmentation $S$ also centered at the provided spatial location, and pair of candidate segment ids $(a,b)$, the two additional channels correspond to binary masks $\mathds{1}(S=a)$ and $\mathds{1}(S=b)$.

\subsection{Additional Information}
\label{sec:addt_info}

First, we explored whether shape information, encoded as point clouds with points randomly sampled from the volumes of the candidate segments to merge, would be helpful as an additional source of information.  Shape information extracted in this manner would have two potential benefits beyond grayscale: First, by using a point cloud representations, a larger context size can be considered than the receptive field available to the CNN.  Second, by reducing reliance on grayscale, such a method may be more amenable to cross-data set generalization, allowing the ground-truth from one reconstruction to be applied to a related reconstruction.  

The specific model we used for binary prediction with a point cloud representation is EdgeConv~\cite{wang2019dynamic}.  We evenly sampled half the points in each candidate segment.  Points were not further distinguished between segments, so prediction was based on the shape of the union of the two segments.

Next, we considered incorporating synapse-level information.  We start with the intuition that segments belonging to the same underlying neuron should likely have similar connectivity patterns, meaning the segments share common inputs and outputs.  We therefore construct feature vectors consisting of synapse counts to the most common inputs and outputs of the pair of candidate segments, and additionally the same information at the cell-type level for inputs/outputs instead of neuron-level.  This feature vector is then used as input to an SVM that is trained using the focused merge data~\cite{cortes1995support}.

Additionally, we further incorporated synapse-level information by adding a fourth channel to the above CNN model, consisting of a binary mask indicating proximity to a predicting synapse.  The intuition for this channel was derived from manual annotation work where it was observed that true termination points of a neuron often were at synapse locations, whereas absence of a synapse at a fragment boundary was often indicative of a false split.

Finally, a currently unused but potentially useful source of information for automated proofreading is neurotransmitter predictions~\cite{eckstein2024neurotransmitter}.  With the underlying assumption of one neurotransmitter type per neuron, neurotransmitter predictions may be a strong signal for whether two fragments should be merged, as well as an indicator for segments that may contain a false merge and should be considered for the cleaving workflow.  We have conducted preliminary experiments using neurotransmitter predictions for identifying false merges, but found a naive search to be overly noisy due to the accuracy level of the neurotransmitter predictions.

\section{Results}

We validate and apply Autoproof to a complete connectome reconstruction of the \emph{Drosophila} male central nervous system (CNS).

\subsection{Data}

The underlying EM grayscale volume was imaged at $8x8x8$ nm resolution, with a total image volume of 160 teravoxels.  Segmentation consisted of roughly 90 million segments, generated using the method of Flood Filling Networks~\cite{januszewski2018high}, and connectivity consisted of 300 million predicted synapses.

The segmentation and other reconstruction data was managed by a Connectomics versioned dataservice, DVID~\cite{katz2019dvid}, that allows retrieval of segmentation state before and after manual proofreading. Since the system is used during manual proofreading workflows as part of our reconstruction efforts, it provided a ready source of information for model training.

\subsection{Focused Merge}

To validate Autoproof, we first used the manual annotations produced by the proofreaders in the focused merge workflow.  These annotations were divided into training and test, with 20k focused merge decisions reserved for test, and 40k decisions used for training.  The training examples were balanced between merge and don't merge decisions, selected from the overall distribution, whereas the test set reflected the global distribution with an overall merge rate of 20\%.

For the CNN model, we used a receptive field size of $130^3$ voxels. For the point cloud representation, points were sampled from a segmentation of size $300^3$ at a downsampled factor of 4 for an effective context size of $1200^3$.  We sampled 2048 points for the point cloud, and set $k=20$ in EdgeConv.  Prediction scores from multiple sources of information (CNN, point cloud, synapse) were combined with an SVM.

\begin{figure}[htbp]
\centering
\includegraphics[width=0.8\textwidth]{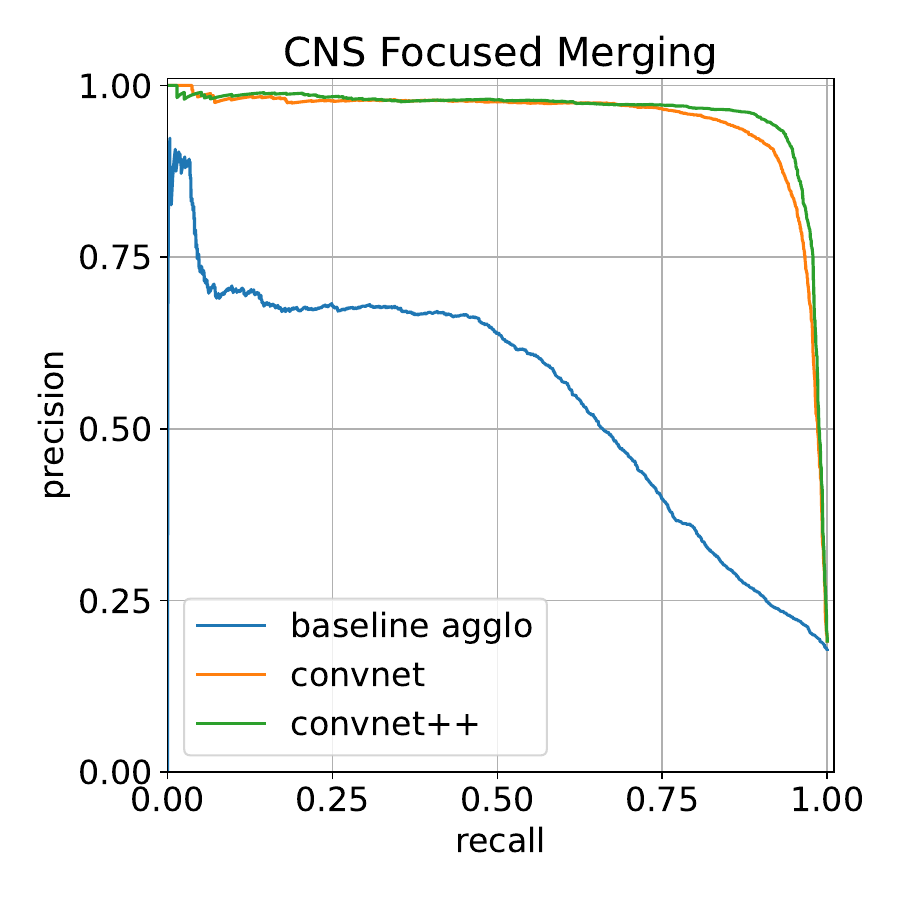}
\caption{Focused merge precision-recall.  Baseline performance using scores from the initial segmentation algorithm is in blue (baseline agglo).  Performance using the CNN is given in orange (convnet), and performance using the CNN combined with shape and synapse information is give in green (convnet++).}
\label{fig:focused_merge}
\end{figure}

Performance on the focused merge test set is given in Figure~\ref{fig:focused_merge}.  As noted earlier, the set of candidate segments for the focused merge protocol is generally produced by the initial segmentation algorithm, in this case the agglomeration step of Flood Filling Networks.  Therefore, a natural baseline for comparison is the precision-recall curve given by directly using the scores produced by the initial segmentation model.  This baseline curve is shown in blue.  

Using only the convolutional network gives the orange curve labeled ``convnet''.  This method, incorporating the ground-truth produced by the manual proofreading (but distinct from the test set), significantly improves upon the baseline.  Finally, using an SVM to combine the predictions from convnet, with the shape-based point-cloud representation and the synaptic connectivity information yields the green curve ``convnet++'', which gives a slight boost beyond ``convnet''.

As can be seen in the figure, Autoproof is able to achieve high precision through a range of recall values, suggesting the possibility of automatically accepting some fraction of the Autoproof decisions.  As a more conservative alternative, the Autoproof scores could also be used to reorder and filter the focused merge tasks that are presented to manual proofreaders.  As Autoproof is able to achieve above 90\% precision at 90\% recall, and the empirical overall merge rate in the focused merge tasks was 20\%, the above results imply that 90\% of the merges (and therefore 90\% of the value) can be recovered in 20\% of the time, with no loss of accuracy.  (90\% precision at 90\% recall implies false positives equal to false negatives, therefore the number of positive predictions equaling the total number of merges.)

\subsection{Orphan Link}

With the above results, we further observe that we can reduce the problem of orphan link to focused merge in the following manner: Given an orphan fragment, if we can generate possible candidate merges, with high recall at possible expense of precision, we can then apply a model trained on focused merge to each generated candidate pair to determine where to connect the orphan fragment.  

A simple approach to generating potential candidate merges given an orphan fragment is to use segmentation spatial adjacency.  We compute a table of all adjacency edges at a downsampled factor of 8, dividing the full segmentation into smaller blocks and computing edges within each block in parallel.  

We first ran a small pilot study of this proposed approach for orphan link by testing the method on ground-truth data from the orphan link workflow.  We found that the method was able to achieve high precision for some range of recall, although lower recall than with the focused merge data.

Given the large number of orphan fragments, it is intractable to examine all such fragments manually in the orphan link workflow.  We used Autoproof to process all unexamined orphan fragments with a synaptic weight between 10 to 100, for approximately 1 million orphan fragments considered.  As most of the predictive value in focused merge came from the CNN model, we chose to apply just the CNN model to orphan link.  As before, we computed spatial adjacency neighbors to generate candidate merges, and examined all edges between an orphan link fragment and a proofread neuron.

To evaluate the performance of the proposed Autoproof orphan link merges, we randomly sampled 2000 merges for manual review.  A first pass was performed where a proofreader marked each merge that they felt was unambiguously correct.  Afterward, the remaining merges were additionally examined by a second proofreader, and both proofreaders were instructed to assess each proposed merge as being correct, incorrect, or indeterminate given the image data.  

\begin{figure}[htbp]
\centering
\includegraphics[width=\textwidth]{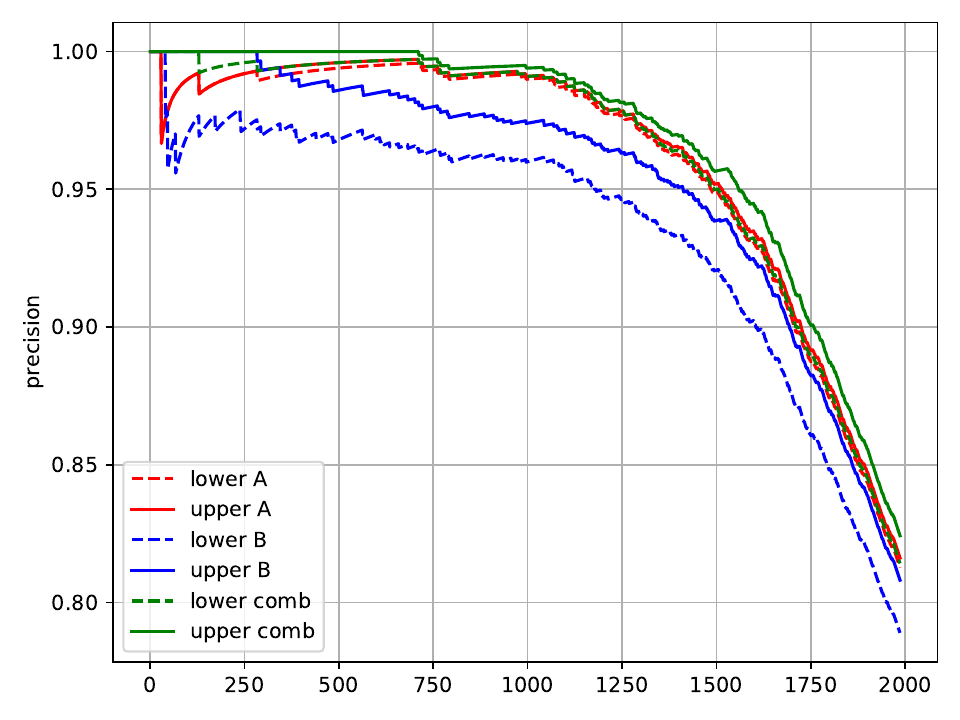}
\caption{Precision of randomly sampled Autoproof proposed orphan link merges.  Dotted curves show precision if all merges marked ``indeterminate'' are considered false, solid curves if considered true.  Green combined curve shows precision if a merge is considered true if either proofreader A or B marked the merge as true.}
\label{fig:orphan_link}
\end{figure}

Figure~\ref{fig:orphan_link} shows plots of precision for the sampled merges.  Each proofreaders' evaluation is plotted twice, once where merges marked indeterminate are considered false (should not merge), and considered true.  The proofreaders' evaluations are also combined, by taking the max of their individual assessments (\eg true if either proofreader marked true).

We observe that there is some variability in manual proofreader assessments; however, overall the Autoproof merges achieve high precision above $0.95$ for much of the sampled range.  From this analysis, we selected a conservative threshold, corresponding to an expected error rate of less than 3\% based on proofreader A's assessments.

By applying Autoproof to the CNS orphan link workflow, we were able to generate and automatically accept 200 thousand merges.  These merges added 309 thousand pre-synaptic T-bars and 2.4 million post-synaptic densities to the reconstruction, thereby increasing the connectivity completion rate by 1.3\% points.  Using an estimate for orphan link proofreader time of about 200 tasks per day, our results of our automated procedure was equivalent to about 4 person-years of work.

\section{Conclusions and Future Work}

We have described Autoproof, a system for automating segmentation proofreading to reduce bottleneck manual annotation costs in EM connectome reconstructions.  Our core idea is to leverage the ground-truth data that is produced in the manual proofreading workflows and that is otherwise not being used in machine learning models.  We first demonstrate that such a system can be trained from focused merge proofreading data, achieving high accuracy on the focused merge workflow and allowing for a 20\% reduction in time while maintaining 90\% of the value in terms of identified merges.  

We then show how a system trained on focused merge can be applied to the unguided orphan link workflow.  We applied Autoproof to 1 million orphan fragments that would otherwise have been unexamined, and were able to generate 200 thousand merge proposals with sufficient accuracy to be automatically accepted.  This procedure, equivalent to 4 person-years of work, was able to increase the connectivity completion rate of the reconstructed connectome by 1.3\% points.

As segmentation algorithms improve, the quantity and quality of errors to manually correct will change.  However, it is unlikely in the near-term that automatic segmentation quality will reach a level where manual proofreading is no longer required.  In addition, the resources needed in terms of computation and available ground-truth to train state-of-the-art segmentation models may not be available to smaller EM reconstruction efforts.  For these reasons, we believe Autoproof will continue to be valuable as a in-the-loop system that is adaptively trained on the available ground-truth data from manual proofreading during a reconstruction.

Additionally, as a next goal, we will examine the problem of cross-data set generalization, to assess whether the ground-truth available from manual proofreading on existing reconstructions can be applied to new, related reconstructions.  This generalization may require making more use of information signals that are more invariant across data sets, such as the shape and connectivity-based features discussed in Section~\ref{sec:addt_info}.  We also believe Autoproof will be amenable to newer imaging modalities such as LICONN~\cite{tavakoli2025light}, and again may be useful in reducing the time and cost needed for such reconstructions.  Finally, we believe that self-supervised learning may provide another rich source of information useful for automated proofreading, as such techniques have been shown useful for problems such as identification of subcompartments and cell types.

\bibliography{auto}
\bibliographystyle{ieeetr}

\end{document}